\documentclass{article}

\PassOptionsToPackage{numbers, compress}{natbib}

\usepackage[preprint]{neurips_2022}

\usepackage[utf8]{inputenc} 
\usepackage[T1]{fontenc}    
\usepackage{hyperref}       
\usepackage{url}            
\usepackage{booktabs}       
\usepackage{amsfonts}       
\usepackage{nicefrac}       
\usepackage{microtype}      
\usepackage{xcolor}         
\usepackage{algorithm}
\usepackage{amsmath}
\usepackage{delimset}
\usepackage{cleveref}
\usepackage{caption}
\usepackage{subcaption}
\usepackage{graphicx}
\usepackage[toc,page]{appendix}
\usepackage{booktabs} 
\usepackage{multirow}
\usepackage{adjustbox}

\usepackage[noend]{algpseudocode}
\newcommand{\RR}[1][]{\mathbb{R}^{#1}}
\newcommand{\EE}[1][]{\mathbb{E}_{#1}}

\title{Imbalanced Classification via Explicit Gradient Learning From Augmented Data}

\author{%
  Bronislav Yasinnik \\
  Department of Computer Science\\
  University of Tel-Aviv\\
  Tel-Aviv, Israel  \\
  \texttt{bronislavy@mail.tau.ac.il} \\
   \And
   Moshe Salhov \\
   Department of Computer Science \\
   University of Tel-Aviv \\
   Tel-Aviv, Israel \\
   \texttt{moshes@playtika.com} \\
   \AND
   Ofir Lindenbaum\\
   Faculty of Engineering, Bar Ilan University \\
   Ramat-Gan, Israel \\
   \texttt{ofirlin@gmail.com} \\
    \AND
   Amir Averbuch\\
   Department of Computer Science\\
  University of Tel-Aviv\\
  Tel-Aviv, Israel  \\
   \texttt{amir1@tauex.tau.ac.il} \\
}

\begin{document}

\maketitle

\begin{abstract}
 Learning from imbalanced tabular data is a significant challenge in real-world classification tasks.
In such cases, neural network performance is substantially impaired due to implicit bias toward the majority class. Existing solutions attempt to eliminate the bias through data re-sampling or re-weighting the loss in the learning process. Still, these methods tend to overfit the minority samples and perform poorly when the structure of the minority class is highly irregular. Here, we propose a novel deep meta-learning technique to augment a given imbalanced dataset with new minority instances. These additional data are incorporated during the classifier's training process, and their contributions are learned explicitly. The augmented samples are modified throughout the training to optimize the classifiers' average-precision score on a validation set. Multiple experiments with synthetic and real-world imbalanced datasets demonstrate the advantage of the proposed method, leading to a significant gap in comparison to many existing baselines.
\end{abstract}

\section{Introduction}
 \par The era of big data has introduced many challenges to machine learning models. In particular, imbalanced classification is a central challenge in a world of exceedingly growing amounts of data. Many practical applications that involve tabular data are trying to learn from imbalanced datasets with extremely skewed class distributions, for example, credit fraud detection in the banking industry \cite{awoyemi2017credit}, rare earthquake identification from seismic data \cite{rabin2016earthquake,bregman2021array}, customers churn prediction in marketing \cite{gordini2017customers}, to name a few. 
 
 \par Neural networks are usually trained on carefully designed datasets where the collected data is labeled and consists of a balanced amount of representative samples from each class. Unfortunately, real-world observations are often severely imbalanced. As a result, most neural networks trained on such datasets tend to be biased towards the majority class while failing to recognize the rare underrepresented class.

\par In order to mitigate this issue, a number of approaches have been proposed to correct data-set imbalance. Usually, these methods concentrate on balancing via re-sampling the training data or re-weighting the objective loss function. However, there are different risks involved in re-sampling. In upsampling, for example, the risk is over-fitting the samples in the minority class, which can lead to learning incorrect information from noise and outliers as these become overrepresented. On the other hand, the under-sampling approach suffers from losing valuable information as too many samples in the majority class which carry useful information are ignored during training.  

\par Other methods try to correct the imbalance by generating new samples in the minority class of the underlying distribution. However, these methods often assume apriori knowledge of the geometric structure of the data manifold. For example, SMOTE \cite{Smote}  generates synthetic samples on the convex neighborhood of each minority instance. The generated points contribute to the classifier when the minority class is well clustered and sufficiently discriminative. However, in real-life datasets, the minority class is often poorly represented and lacks a clear structure. 

\par  Here, we propose a novel meta-learning framework to augment a given imbalanced dataset with new minority instances. After the neural network classifier is initially trained over the original dataset, the minority class samples in regions of high uncertainty are duplicated. These duplicates are then incorporated in the classifier's learning process, and their position is iteratively optimized in a two-steps routine: 
first, we train on a batch of input data that consists of the new instances. During the backpropagation of this step, we explicitly calculate the gradient update of each network parameter as a function of the input samples. During the second step, we examine this gradient update by passing a validation set through the updated network and calculating the loss. In the final step, we backpropagate through this loss to modify the input samples' positions so that they are aligned to create a better minority class distribution boundary.

\par Our main contributions are summarized below:
\begin{itemize}
    \item We show that it is possible to produce suitable synthetic samples in the minority distribution by optimizing an estimator of the classification performance.
    \item  We demonstrate that the obtained samples allow neural net classifiers to improve classification objectives when the distribution of class samples is highly imbalanced.  
    \item Our method outperforms traditional re-sampling methods on many real-world tabular datasets. Additionally, it has the advantage that it can be learned in an end-to-end fashion, simultaneously optimizing both the classifier and the synthetic data.
    
\end{itemize}

\section{Related Work}
There are several approaches for dealing with imbalanced data in the literature. These approaches can be roughly divided into two categories, namely data-level methods, algorithm-level methods.

\par \textbf{Data-level methods.} These methods mainly focus on data re-sampling strategies to improve classification; the two primary ways are over-sampling and under-sampling. In up-sampling, we create a balanced data-set by repeating minority examples until their number reaches the size of the majority class. In under-sampling, we reduce the size of the majority class by removing random majority examples until their number reaches the size of the minority class. There are also more advanced approaches that attempt to balance classes' size by generating synthetic data. For instance, SMOTE \cite{Smote} produces new minority samples by interpolating between existing minority samples and their nearest minority neighbors. SMOTE has some improved variants attempting to generate better quality samples. Borderline-SMOTE \cite{BorderlineSmote} tries to limit the generated samples to be near the class borders,  AdaSyn \cite{Adasyn} adaptively generates ``harder'' minority examples according to their density in the regions of the feature space. Another approach titled SUGAR \cite{Sugar}, aims at generating points uniformly along the manifold of the data.

\par \textbf{Algorithm-level methods.} In contrast to data-level methods, these methods do not alter training data distribution. Instead, they employ class-sensitive losses, i.e., each class is associated with a different weight. For example, increasing the weight of the minority class reflects the increase of its importance. In the simplest form, the weights are assigned inversely proportional to the population size of the class. During training, the losses incurred are re-weighted according to the assigned weights, e.g., class balanced loss \cite{cui2019class}. More carefully designed losses include Focal Loss \cite{FocalLoss}  which puts more significant weight on ``harder'' examples and its extension Asymmetric Focal Loss \cite{AsymmetricFocalLoss} that has an additional parameter that decouples the focusing levels of the positive and negative samples. Other common strategies are thresholding methods that vary the classifier's decision threshold to eliminate the bias towards the majority class. 


\par \textbf{Meta learning methods.} Preliminary efforts in meta-learning on the imbalanced classification task such as Meta-Weight-Net \cite{metaweightnet},  are attempting to learn a mapping from data space into the classifier weights space by learning explicit weighting function measuring the importance of each sample. Similarly, MESA \cite{mesa} learns how to resample the data via ``meta-sampler" through a reinforcement learning approach. Our method is different since we are not trying to learn a resampling strategy of the existing data. Instead, we are using meta-learning in order to learn how to create the new samples dynamically and improve the classifier performance by augmenting the imbalanced dataset with new samples.

\section{The proposed method of Augmentation by Explicit Gradient Meta-Learning}
\par In a binary imbalanced classification task. Let's denote the training dataset $D = (X, Y)$ and the neural network classification model $C_{\theta}$ where $\theta \in \RR[k]$ are the network parameters. Since we only have access to a small amount of samples from the distribution of the minority class, and we need to neutralize the dominance of the majority samples during training, we propose to augment the minority class with synthetic samples $D^{synth} =(X^{synth}, Y^{synth})$. With the new $D^{synth}$, we will create a modified classifier that will generalize better to unseen data at test time (see \cref{algorithm:meta}). 

\par Generate informative synthetic samples for tabular data is a challenging task \cite{challenges}, especially when the minority class is very small. Instead, we suggest to duplicate the existing minority samples which we call $D^{synth}$ and  training the classifier on a dynamically changing $X^{synth}$ using a meta-learning framework. A key idea behind this method is that after introducing the new synthetic samples, the classifier learn the explicit gradient that should be applied directly on these data-points in order to adjust the minority region in feature space.  Consequently, by tunning $C_{\theta}$ on both $X^{train}$ and $X^{synth}$, we can determine a more accurate decision boundary which is less biased toward the majority class.  

\par Our implementation requires an initial classifier that will be modified gradually through the introduction of samples in $D^{synth}$. We will split $D$ into two disjoint subsets $D = D^{train} \sqcup D^{valid}$ each of which is distributed as the original $D$. The first subset, $D^{train}$, would serve us to obtain an initial classifier $C_{\theta_0}$. The second subset, $D^{valid}$, controls how $C_{\theta}$ adapts to the synthetic samples while remaining unbiased with respect to the original data.

\par We aim to enhance performance in the challenging regions of feature space. To do that, we obtain synthetic samples $D^{synth}$ by duplicating the minority samples that lie in regions of high uncertainty as determined by $C_{\theta_0}$, i.e., we choose all minority samples $x\in D^{train}$ for which $C_{\theta_0}(x)\leq c$, for some predefined constant $0\leq c\leq 1$. In practice in all our experiments the datasets are relatively small and the imposed imbalance ratio only leaves few minority points, we found that it is more effective to set $c=1$ and duplicate all the minority data. In larger datasets where the imbalance ratio is less severe it is worth to tune the parameter $c$ to attain better performance.

\par The training of our classification model $C_{\theta}$ involves the following steps: First  we adapt the decision boundary of $C_{\theta}$ in terms of the synthetic data. We do this by calculating the loss function on $X^{synth}$ and performing one or more gradient descent updates.  We call the resulting classifier parametrized by  $\theta'(X^{synth})$  the ``adapted classifier''. Note, the new parameters are calculated as  a function of $X^{synth}$, for example using one gradient update:
\begin{equation}
    \label{eq:1}
     \theta'(X^{synth}) = \theta - \eta_1 \nabla_{\theta} \mathcal{L} (X^{synth}; C_{\theta}).
\end{equation}

\par Next, we gradually modify $X^{synth}$. This is achieved while preserving the ``adapted classifier'' fidelity to the original data.  For this purpose, we compute the loss of the adapted classifier on the validation set, i.e, $\mathcal{L}(D^{valid}; C_{\theta'})$. We change the value of $X^{synth}$, by performing gradient descent through the computed loss, i.e.: 
\begin{equation}
    \label{eq:2}
    X^{synth} \gets X^{synth} - \eta_2 \nabla_{X^{synth}}\mathcal{L}(D^{valid}; C_{\theta'}).
\end{equation}

With our explicit gradient method, we optimize the classifier parameters by performing one or a small number of gradient steps on $Z^{synth}$, and then move them to a region that produces a better final decision boundary. The total objective can be formulated as follows:
\begin{align*}
\min_{\theta}\min_{X^{synth}}\EE\brk[s]{\mathcal{L}(D^{train}, X^{synth};C_{\theta})} = \\  =\min_{X^{synth}}\EE\brk[s]{\mathcal{L}(D^{train};C_{\theta - \alpha \nabla_{\theta} \mathcal{L} (X^{synth}})}.
\end{align*}
The update on $Z^{synth}$ involves a gradient through gradient which requires computation of the Hessian vector products and is supported by standard deep learning libraries such as Pytorch \cite{PyTorch}.
\begin{algorithm}
\caption{ Meta Learning  on Synthetic Data}\label{algorithm:meta}
\begin{algorithmic}[0]
\State \textbf{Input:} 
\State Training dataset: $D=(X, Y) $,  $ X \subseteq \RR[d] ,  Y \subseteq {\{0,1\}}^n $.
\State Classifier $C_{\theta}: \mathbb{R}^d \rightarrow \{0,1\}$.
\State \textbf{Initialization:}
\State 1. Split $D$ into disjoint training and validation sets having the same class distribution as $D$:
\[
D = D^{train} \sqcup D^{valid}.
\]
\vspace{0.1cm}
\State 2. Train a classifier only on $D^{train}$ and set the weights for  $C_{\theta_0}$
\vspace{0.1cm}
\State 3. Initialize the synthetic set by duplicating the minority instances in $D^{train}$, for which $C_{\theta_0}$ is least confident up to a confidence bound  $c\in [0,1]$ :
\[
D^{synth} = \brk[c]{(x,y)\in D^{train}\mid y=1, C_{\theta_0}(x) \leq c}.
\]

\vspace{0.1cm}

\For{ minibatches $(X_1, X_2, X^{synth}) \subseteq D^{train} \times D^{valid} \times D^{synth}$}
\State \textbf{step 1:} 
\State Calculate the first order update of the model as function of the synthetic batch $X^{synth}$:
\State \hspace{4cm} $\theta(X^{synth}) \gets \theta-\eta_1\nabla_{\theta}\mathcal{L}(X_1\cup X^{synth}\mid \theta).$ \vspace{0.2cm}


\State \textbf{step 2:} 
\State Update $X^{synth}$:
\State \hspace{4cm} $X^{synth} \gets  X^{synth}-\eta_2\nabla_{X^{synth}}\mathcal{L}( X_2 \mid \theta(X^{synth})).$

\EndFor
\end{algorithmic}
\end{algorithm}

\section{Imbalanced Regression on Synthetic Dataset}

\begin{figure*}[ht]
\vskip 0.2in
\begin{center}
\centerline{\includegraphics[width=\textwidth, height=5cm]{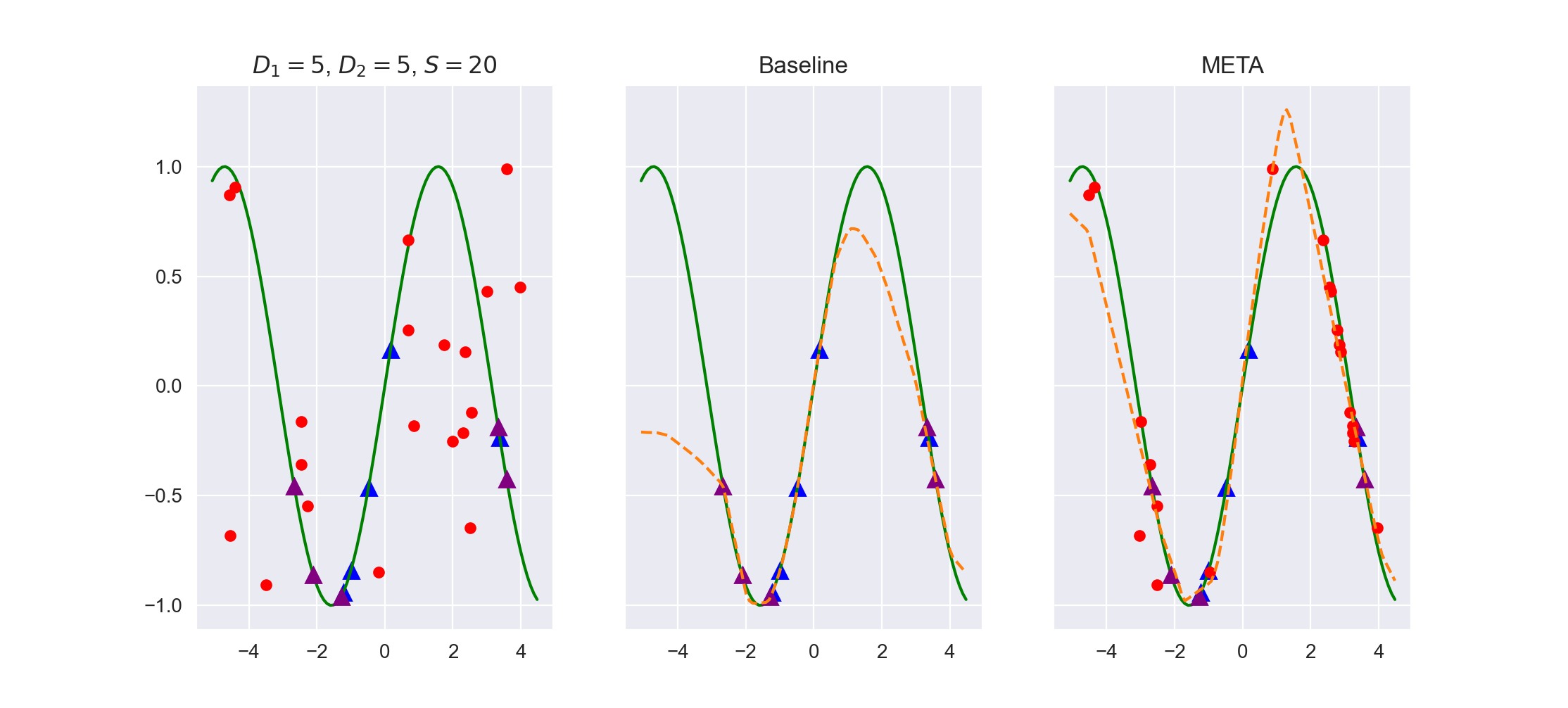}}
\caption{
Illustration of imbalanced regression using our explicit gradient method to learn the true sine wave (green line) via  adaptation on randomly generated augmentations. \textbf{Left:} The dataset consists of $10$ samples on the sin wave; we partitioned them into 5 points (blue triangles) for the training set and 5 (purple triangles) for the validation set. The generated points (red) are sampled uniformly in the span $[-1,1]$ here we use 20 new samples. \textbf{Middle:} Training a neural-net regression model (dashed orange) to fit the actual data. \textbf{Right:} The resulting model after meta-training on the synthetic samples. The final model fits the regression task better, and the synthetic points are aligned close to the objective sine function. 
}
\label{fig:sin}
\end{center}
\vskip -0.2in
\end{figure*}

\par In order to demonstrate how our approach works, we present how we utilized our meta-learning algorithm to improve upon an imbalanced regression task by generating auxiliary synthetic samples. We show how our method enables a neural network to learn a regression of a sine wave function by training only on a few real input points.
\par In this example, the true function is a sine wave with amplitude $1$, frequency $\pi/2$, and phase $0$. We use a set of $D_1=5$ datapoints for $X^{train}$, a set of $D_2=5$ points for $X^{valid}$ also we supply $S=20$ randomly generated points as $X^{synth}$ (\cref{fig:sin} Left). All the input points are sampled uniformly from $\brk[s]{-5.0, 5.0}$. The loss is the mean-square error between the prediction $f(x)$ and the true value. The regressor is a neural network model with 1 hidden layer of size 10 with Tanh nonlinearities. The step size we have set for \ref{eq:1} is $\eta_1=0.1$ and for \ref{eq:2} is $\eta_2=1$. 
\par For a baseline, we train the regressor on both $D^{train}\cup D^{valid}$ (\cref{fig:sin} Middle). Using our method we first train the network only on $D^{train}$ and then fine-tune via meta-learning on the synthetic points by  $D^{valid}$. In \cref{fig:sin} we see how our meta-learning approach succeeds to shift the data points closer to the objective sine function by learning the gradient of the added data points explicitly and encouraging the adapted model to improve the loss on the validation set. In \cref{appendix:a}, we compare additional scenarios when we vary the number of synthetic points and real points. 

\section{Experiment on Induced Imbalance}

\par \textbf{Setup for induced imbalance:} We create different levels of artificial imbalance on initially balanced
data by dropping instances from the minority class according to the following increasing target ratios $50:1$, $100:1$ and $200:1$. On each of the imbalanced versions of the data, we train the same neural network using several imbalance learning benchmarks. We compare to a ``potential" baseline in order to assess the resulting gap in performance to a perfect method which is trained after imputing some of the withdrawn samples and doubling the size of the minority class in the dataset. Our goal here is to quantify the gap that separates our approach from an ideal method that can reproduce synthetic points where real unseen minority samples reside.

\par \textbf{Dataset and prepossessing:} We carry out this experiment on \textbf{adult} \cite{adult}. The data contains census information about adult citizens, and the task is a binary classification where the goal is to predict if the income of a citizen exceeds $50K\$/year$. There are two target classes labeled $>50K, <=50K$, where the minority class is labeled by  $>50K$. The data comprises of 14 features, eight categorical features, and six continuous features. Initially, the total number of examples was $48842$. To create a balanced dataset, we drop the instances containing missing features and randomly remove samples from the majority class until both classes are equal in size. The remaining dataset is balanced and contains $22416$ instances in total. For the data preprocessing, we apply normalization to the continuous features and one-hot encoding for the categorical features. We reduce the dimension of the processed data using PCA from $101$ to $64$.


\par \textbf{Baselines:} We compare our proposed method with widely adopted baselines that are used for training on imbalanced data.  (1) \textbf{CE} - Training without any special treatment by using the standard cross-entropy loss  (2) \textbf{RS} - re-sampling such that each minibatch contains the same number of representatives of each class (3) \textbf{SMOTE}- doubling the size of the minority class by generating samples in their neighborhood (4) \textbf{Focal}- we use the focal-loss \cite{FocalLoss} as representative for the re-weighting methods.

\par \textbf{Model:}  The model we use is a standard MLP classifier consisting of two hidden layers of size $[128, 16]$ and RELU activation.  For training the baselines we use Adam \cite{Adam} with learning rate $10^{-3}$ and $\beta_1=0.5, \beta_2=0.999$. The model is trained with a batch size of 128 for 200 epochs. For the meta-learning we use $\eta_1=10^{-2}, \eta_2=10$, the synthetic data is created by duplicating all the minority samples by setting $c=1$. In our experiments we use a stratified 4-fold train/test split with $80\%$-$20\%$ ratio and $20\%$ from the train set as validation.

\par
\begin{figure*}[ht!]
    \centering
    \includegraphics[width=0.3\textwidth, height=5cm]{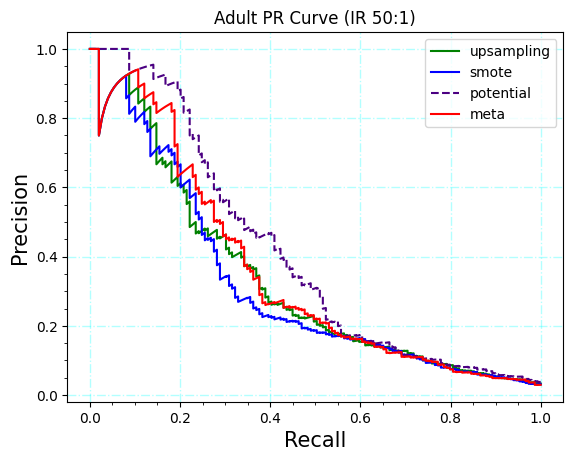}
    \includegraphics[width=0.3\textwidth, height=5cm]{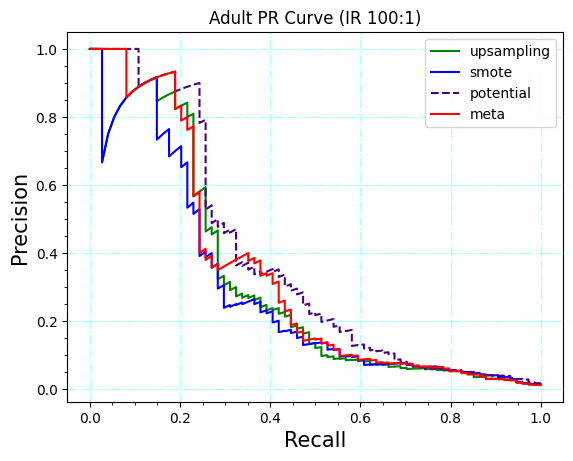}
     \includegraphics[width=0.3\textwidth, height=5cm]{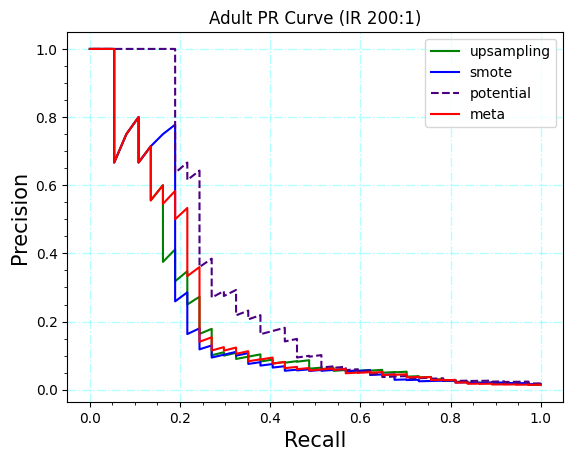}

    \caption{Comparing precision-recall of the model trained on adult training set with different imbalance ratios \textbf{Left}: (50:1) \textbf{Middle}: (100:1), \textbf{Right} (200:1), the depicted curves obtained through the evaluation on the test sets. The compared baselines are Upsampling (green), SMOTE (blue), Ours (red). Our method outperforms all the tested baselines over a wide range of recall values. The potential (dashed purple) depict the possible lift in performance if instead of the synthetic points we would balance the data using additional minority samples.  }
    \label{fig:PR-adult}
\end{figure*}

 \textbf{Results:} In all our experiments with the induced imbalance on the \textbf{adult} dataset, we have demonstrated that with varying imbalance ratios our method was able to achieve the best scores. In particular, we showed that our method is advantageous even in extreme imbalances scenarios such as when the imbalance ratio reaches to $200:1$ (see \cref{table:adult_ap}). The $\Delta$ gap in the average precision between our method and the best in potential is the lowest among the compared baselines. In \cref{fig:PR-adult} we observe that our method prevails over a substantial range of recall rates. We also supply a convergence verification of our trained model and the synthetic points see \ref{appendix:c} . 

\begin{table}[ht!]\centering
\caption{Classification results on the adult data set compering our meta classification  with the potential and re-sampling baselines}
\label{table:adult_ap}
\resizebox{0.7\textwidth}{4cm}{
\begin{tabular}{@{}lllllll@{}}
\toprule
\multicolumn{1}{c}{\multirow{2}{*}{\textbf{Method}}} & \multicolumn{1}{c}{\multirow{2}{*}{\textbf{IR}}} & \multicolumn{4}{c}{\textbf{Metrics}} \\ \cmidrule(l){3-7} 
\multicolumn{1}{c}{} & \multicolumn{1}{c}{} & AUC-PR & $\Delta$ & AP@0.75 & AP@0.5 & AP@0.25 \\ 
\midrule
\multicolumn{1}{c}{
    \textbf{CE}} & 50:1 & 0.333 & -18.1\% &\textbf{0.10} & 0.193 & 0.513 \\ 
    \cmidrule(lr){2-2}
     & 100:1 & 0.308 & -17.8\% &0.059 & \textbf{0.164} & 0.441  \\ 
     \cmidrule(lr){2-2}
     & 200:1 & 0.204 & -32.0\% &0.029 & 0.087 & \textbf{0.312} \\ 
     \midrule
\multicolumn{1}{c}{
    \textbf{RS}} & 50:1 & 0.328 & -19.4\% & 0.099 & 0.214 & 0.481 \\ 
    \cmidrule(lr){2-2}
     & 100:1 & 0.311 & -17.0\% & 0.059 & 0.135 & \textbf{0.594}\\ 
     \cmidrule(lr){2-2}
     & 200:1 & 0.204 & -32.0\% & 0.037 & \textbf{0.065} & 0.178 \\
\midrule
\multicolumn{1}{c}{
   \textbf{SMOTE}} & 50:1 & 0.312 & -23.3\% & 0.093 & 0.187 & 0.469  \\
    \cmidrule(lr){2-2}
     & 100:1 & 0.289 & -22.9\% & 0.066 & 0.136 & 0.404  \\
     \cmidrule(lr){2-2}
     & 200:1 & 0.208  & -30.6\% & 0.025 & 0.057 & 0.129 \\ 
 \midrule
 \multicolumn{1}{c}{
    \textbf{Focal}} & 50:1 & 0.301 & -26.0\% & 0.083 & 0.175 & 0.480  \\
    \cmidrule(lr){2-2}
     & 100:1 & 0.299 & -20.2\% & 0.05 & 0.126 & 0.504  \\
     \cmidrule(lr){2-2}
     & 200:1 & 0.211  & -29.6\% & 0.036 & 0.043 & 0.121 \\ 
 \midrule
\multicolumn{1}{c}{
    \textbf{Explicit Gradient(Ours)}} & 50:1 & \textbf{0.348}  & -14.4\% & 0.099 & \textbf{0.221} & \textbf{0.558} \\ 
     \cmidrule(lr){2-2}
     & 100:1 & \textbf{0.333} & -11.2\% & \textbf{0.067} & 0.1491 & 0.413 \\ 
     \cmidrule(lr){2-2}
     & 200:1 & \textbf{0.214} & -28.6\%  & \textbf{0.037} & 0.059 & 0.1538  \\ 
\midrule
\midrule
\multicolumn{1}{c}{
    \textbf{Potential}} & 50:1 & 0.407 & baseline  & 0.106 & 0.310 & 0.678  \\ 
    \cmidrule(lr){2-2}
     & 100:1 & 0.375 & baseline  & 0.061 & 0.226 & 0.678 \\ 
     \cmidrule(lr){2-2}
     & 200:1 & 0.300 & baseline & 0.035 & 0.102 & 0.678 \\
 \bottomrule
\end{tabular}
}

\end{table}

  

\section{Experiments on Real World Data}

\par \textbf{Datasets:} 
In order to verify the effectiveness of our approach, we conducted a comprehensive set of experiments on 27 tabular datasets available as part of the Imbalanced-learn package \cite{imblearn}. The datasets provided here were obtained from various domains, with varying numbers of samples, feature dimensions, and imbalance ratios. The categorical features in these datasets are already presented in numerical form. However, no information is available describing how the values were encoded. In addition, we used min-max scaling to bring all the features to the $[0,1]$ interval. The largest dataset is protein\_homo  which contains 145,751 samples, and the smallest dataset is ecoli, with 336 samples. The number of features ranges between 6 features in mammography and 617 features in isolet. The imbalance ratio ranges between 8.6:1 in ecoli and 130:1 in abalone19. The full description of the datasets is presented in \ref{appendix:b}.

\par \textbf{Evaluation protocol and scores:}
We evaluate the performance of our trained classifier using the traditional classification metrics which include the area under the precision-recall curve (AUC-PR), and the area under the receiving operator curve (AUC-ROC) following the evaluation metrics used in \cite{WGAN} \cite{mesa}. For the experiments that we run we used 4-fold stratified splits as in \cite{mesa} in which we keep out 20\% for the test-set and 60\%/20\% for training and validation. In each of the methods, we also report the SD of scores across the runs. 

\par \textbf{Baseline methods:}
For the \textbf{first set} of experiments, we compare our method with the classical literature baselines that also employ a generation of new minority samples to reduce the class imbalance. We used the same number of generated samples in all experiments to ensure a fair comparison. The compared methods are SMOTE \cite{Smote}, AdaSyn \cite{Adasyn}, and BorderlineSmote \cite{BorderlineSmote} ; we used the implementations of these methods in the Imblearn package \cite{imblearn}. Additionally, we include the results when the classifier is trained with the original data when no resampling is applied and minimizing the cross-entropy loss  (CE) and the results when the minority class is randomly oversampled (ROS) to the same ratio as the other baselines. We report the results for this set of experiments in \ref{table:results_variants_smote}.

\par For the \textbf{second set} of experiments, we compared our results with recent deep-learning methods for imbalance correction. We divide these methods into three categories and present a comparison of several candidates in each category. In the first category of loss-based methods, we include Focal Loss \cite{FocalLoss} and LDAM loss \cite{LDAM} as representatives. The second category we compare is generative methods. Specifically, we found cWGAN \cite{WGAN} as the most suitable candidate since the generator and discriminator networks are designed ad hoc for tabular data. The final classification is done by the same classifier architecture as was used in the previous experiments. We also note that there are plenty of generative methods for dealing with the imbalanced data problem in the literature such as AC-GAN \cite{ACGAN}, D2GAN \cite{D2GAN}, DeepSmote \cite{DeepSmote} however their main consideration is the image domain and it is not clear if they can be easily adapted to the tabular domain. To get the results on our datasets we rerun the cWGAN backbone architecture with changes tailored to our data. In particular, we keep only the layers that treat numerical features while disconnecting the layers responsible for categorical features both in the generator and discriminator. This can be easily configured by setting zero categorical features in the model configuration. We used the same values for the hyperparameters in cWGAN as the authors suggested for their experiments. Lastly, we compare a meta-learning method called MESA \cite{mesa} which trains a meta-sampler through reinforcement learning to know how to resample the data during classifier training. We report the AUC-PR metric only on the 5 datasets which were provided by the authors.

\begin{table}[]
\caption{AUC-PR on the Imbalanced-learn package benchmark datasets. The comparison is between our method and the traditional methods the reported score is the mean and std over 4-folds run. }
\label{table:results_variants_smote}
\centering

\resizebox{0.8\textwidth}{5cm}{
\begin{tabular}{@{}lllllll@{}}
\toprule
Method           & CE        & ROS       & SMOTE     & ADASYN    & BSMOTE    & Ours \\ \midrule
ecoli            & 0.64+0.16 & 0.72+0.18 & 0.71+0.17 & 0.68+0.18 & 0.72+0.19 &  \textbf{0.78+0.01}     \\
optical\_digits  & \textbf{0.99+0.01} & \textbf{0.99+0.01} & \textbf{0.99+0.01} & \textbf{0.99+0.01} & \textbf{0.99+0.01} &  \textbf{0.99+0.01}     \\
satimage         & 0.60+0.03 & 0.67+0.04 & 0.70+0.02 & 0.63+0.03 & 0.59+0.03 & \textbf{ 0.72+0.00 }    \\
pen\_digits      & \textbf{1.00+0.00} & \textbf{1.00+0.00} & \textbf{1.00+0.00} & \textbf{1.00+0.00} & \textbf{1.00+0.00} &  0.99+0.00     \\
abalone          & 0.34+0.02 & 0.33+0.01 & 0.33+0.02 & 0.33+0.03 & 0.32+0.02 &  \textbf{0.38+0.01}     \\
sick\_euthyroid  & 0.79+0.03 & 0.78+0.02 & 0.77+0.04 & 0.75+0.02 & 0.79+0.02 &  \textbf{0.82+0.01}     \\
spectrometer     & 0.92+0.04 & 0.91+0.07 & 0.88+0.08 & 0.91+0.06 & 0.92+0.06 &  \textbf{0.96+0.01}     \\
car\_eval\_34    & \textbf{1.00+0.00} & \textbf{1.00+0.00} & \textbf{1.00+0.00} & \textbf{1.00+0.00} & \textbf{1.00+0.00} &  0.99+0.01     \\
isolet           & 0.94+0.01 & 0.91+0.06 & 0.93+0.00 & 0.91+0.04 & 0.74+0.27 &  \textbf{0.96+0.01}     \\
us\_crime        & \textbf{0.55+0.06} & 0.48+0.03 & 0.50+0.07 & 0.45+0.05 & 0.52+0.04 &  0.53+0.00     \\
yeast\_ml8       & 0.11+0.01 & 0.10+0.01 & 0.10+0.02 & 0.10+0.02 & 0.11+0.02 &  \textbf{0.14+0.00}     \\
scene            & 0.19+0.07 & 0.19+0.02 & 0.20+0.05 & 0.18+0.03 & 0.20+0.02 &  \textbf{0.25+0.00}     \\
libras\_move     & 0.86+0.08 & 0.86+0.06 & 0.85+0.08 & 0.86+0.08 & 0.86+0.08 &  \textbf{0.92+0.01}     \\
thyroid\_sick    & 0.77+0.04 & 0.74+0.08 & 0.71+0.07 & 0.72+0.05 & 0.74+0.07 &  \textbf{0.81+0.02 }    \\
coil\_2000       & 0.14+0.02 & 0.11+0.02 & 0.12+0.01 & 0.11+0.01 & 0.11+0.01 &  \textbf{0.16+0.01}     \\
arrhythmia       & 0.26+0.11 & 0.30+0.13 & 0.29+0.13 & 0.27+0.13 & 0.24+0.11 &  \textbf{0.35+0.01}     \\
solar\_flare\_m0 & 0.12+0.02 & 0.12+0.02 & 0.11+0.01 & 0.11+0.03 & 0.09+0.01 &  \textbf{0.16+0.01}     \\
oil              & 0.56+0.21 & 0.60+0.15 & 0.57+0.21 & 0.56+0.21 & 0.58+0.22 &  \textbf{0.63+0.00}     \\
car\_eval\_4     & 0.98+0.03 & \textbf{0.99+0.02} & \textbf{0.99+0.02} & \textbf{0.99+0.02} & \textbf{0.99+0.02} &  \textbf{0.99+0.00}     \\
wine\_quality    & \textbf{0.29+0.05} & 0.25+0.03 & 0.21+0.03 & 0.21+0.04 & 0.24+0.06 &  \textbf{0.29+0.00}     \\
letter\_img      & 0.99+0.00 & \textbf{1.00+0.00} & 0.99+0.01 & 0.99+0.00 & 0.97+0.02 &  0.99+0.01     \\
yeast\_me2       & \textbf{0.48+0.10} & 0.32+0.11 & 0.36+0.15 & 0.33+0.12 & 0.37+0.14 &  0.36+0.01     \\
webpage          & 0.78+0.03 & 0.79+0.01 & 0.79+0.02 & 0.77+0.02 & 0.77+0.02 &  \textbf{0.85+0.00}     \\
ozone\_level     & \textbf{0.31+0.07} & 0.26+0.04 & 0.25+0.02 & \textbf{0.31+0.06} & 0.28+0.03 &  0.30+0.00     \\
mammography      & 0.72+0.05 & 0.71+0.05 & 0.73+0.06 & 0.67+0.08 & 0.68+0.08 &  \textbf{0.75+0.01}     \\
protein\_homo    & 0.86+0.01 & 0.86+0.02 & 0.87+0.01 & 0.86+0.01 & 0.85+0.02 &  \textbf{0.91+0.01}     \\
abalone\_19      & 0.03+0.00 & 0.03+0.01 & 0.03+0.01 & 0.04+0.02 & \textbf{0.06+0.06} &  \textbf{0.06+0.01 }    \\ \midrule
mean             & 0.6+0.33  &   0.59+0.34       &    0.59+0.34       &  0.58+0.33         &   0.58+0.33      &     0.63+0.33   \\
mean rank (out of 10)        &  4.37         &     4.89      &  5.09         &   6.04        &    5.65       &  2.54    \\
\bottomrule
\end{tabular}

}
\end{table}

\begin{table}[]
\caption{AUC-PR scores comparison between our method and contemporary deep-learning techniques from three main categories of imbalanced methods generative (CWGAN), weighted loss (FOCAL, LDAM), meta-learning (MESA).}
\label{table:results_deep}
\centering

\resizebox{0.8\textwidth}{5cm}{
\begin{tabular}{@{}lllllll@{}}
\toprule
Method           & CWGAN                 & FOCAL                & LDAM              & MESA           & Ours               \\ \midrule
ecoli            & 0.76+0.04 & \textbf{0.81+0.00} & 0.69+0.00 &  N/A   &0.78+0.01 \\
optical\_digits  & \textbf{0.99+0.03} & 0.97+0.01 & \textbf{0.99+0.01} &  0.98+0.00  &\textbf{0.99+0.01} \\
satimage         & 0.69+0.03 & 0.66+0.00 & 0.51+0.00 &  N/A      &\textbf{0.72+0.00} \\
pen\_digits      & \textbf{0.99+0.01} & 0.95+0.00 & \textbf{0.99+0.01} &   N/A     &\textbf{0.99+0.00} \\
abalone          & \textbf{0.38+0.16} & \textbf{0.38+0.01} & 0.20+0.01 &   N/A     &\textbf{0.38+0.01} \\
sick\_euthyroid  & 0.74+0.03 & 0.67+0.00 & 0.72+0.00 &   N/A     &\textbf{0.82+0.01} \\
spectrometer     & 0.93+0.04 & \textbf{0.99+0.02} & 0.96+0.01 &  0.84+0.00  &0.96+0.01 \\
car\_eval\_34    & 0.98+0.01 & 0.97+0.00 & 0.91+0.02 &  N/A      &\textbf{0.99+0.01} \\
isolet           & 0.90+0.10 & 0.94+0.02 & 0.84+0.01 &  0.92  &\textbf{0.96+0.01} \\
us\_crime        & 0.48+0.03 & 0.48+0.00 & 0.42+0.02 &  N/A     &\textbf{0.53+0.00} \\
yeast\_ml8       & 0.10+0.05 & \textbf{0.19+0.00} & 0.16+0.01 &  N/A   &0.14+0.00 \\
scene            & 0.23+0.04 & 0.10+0.00 & \textbf{0.39+0.03} &  N/A   &0.25+0.00 \\
libras\_move     & 0.87+0.03 & 0.85+0.01 & 0.87+0.00 &  N/A   &\textbf{0.92+0.01} \\
thyroid\_sick    & 0.74+0.01 & 0.73+0.02 & 0.66+0.01 &  N/A   &\textbf{0.81+0.02} \\
coil\_2000       & 0.12+0.03 & \textbf{0.18+0.00} & 0.17+0.02 &  N/A     &0.16+0.01 \\
arrhythmia       & 0.30+0.02 & 0.26+0.02 & 0.26+0.01 &  N/A     &\textbf{0.35+0.01} \\
solar\_flare\_m0 & 0.15+0.07 & 0.12+0.00 & 0.13+0.00 &   N/A  &\textbf{0.16+0.01} \\
oil              & 0.58+0.06 & 0.56+0.01 & \textbf{0.70+0.02} &   N/A &0.63+0.00 \\
car\_eval\_4     & \textbf{0.99+0.08} & \textbf{0.99+0.01} & 0.94+0.00 &   N/A     &\textbf{0.99+0.00} \\
wine\_quality    & 0.21+0.04 & \textbf{0.29+0.00} & 0.26+0.01 &   N/A     &\textbf{0.29+0.00} \\
letter\_img      & 0.91+0.12 & 0.95+0.00 & 0.94+0.00 &   N/A     &\textbf{0.99+0.01} \\
yeast\_me2       & 0.33+0.00 & \textbf{0.38+0.01} & 0.31+0.02 &   N/A     &0.36+0.01 \\
webpage          & 0.81+0.08 & \textbf{0.85+0.00} & 0.82+0.01 &    N/A    &\textbf{0.85+0.00} \\
ozone\_level     & 0.23+0.01 & \textbf{0.30+0.02} & 0.20+0.00 &   N/A     &\textbf{0.30+0.00} \\
mammography      & \textbf{0.77+0.01} & 0.76+0.01 & 0.69+0.00 &  0.71+0.00  &0.75+0.01 \\
protein\_homo    & 0.81+0.03 & 0.84+0.00 & 0.85+0.01 &  0.86+0.00  &\textbf{0.91+0.01} \\
abalone\_19      & 0.02+0.03 & 0.03+0.00 & 0.01+0.01 &   N/A     &\textbf{0.06+0.01} \\
\midrule
mean             &   0.59+0.33                 &     0.60+0.33               &     0.58+0.32              &     N/A            &  0.63+0.33                \\
mean rank  (out of 10)      &     5.44               &          4.87          &        6.11            &        N/A           &         2.54           \\ 
\bottomrule
\end{tabular}}
\end{table}

\par \textbf{Results:}   \cref{table:results_variants_smote} shows the results of the AUC-PR we obtained on the \textbf{first set} of experiments which include the traditional oversampling and generating methods, we see that our method outperform the other methods on the majority of the datasets. 
In the two largest datasets protein\_homo and webpage we see a significant improvment between $5\%-7\%$. On abalone\_19 where the imbalance ratio is the most severe we can see that we were able to achieve the best score together with BSMOTE and it is $50\%$ higher than the second best. Between the traditional methods there is no particular method that overtake the others. We applied the Friedman test with Iman-Davenport correction as suggested in \cite{WGAN} to the ranks of the methods and the resulting p-value is less than $0.05$ which indicate that the improvement is statistically significant.

\par  In \cref{table:results_deep} we show our results on the \textbf{second set} of experiments comparing to the deep-learning methods, it is also evident here that our method outperforms on most of the datasets. Here we see that between the baselines FOCAL loss has slight advantage. We also see that our method achieves better results than MESA on each of the datasets. Although both methods employ different base classifiers i.e. we use a neural network and MESA decision tree ensembles. In the challenging protein\_homo for example MESA achieves the highest score between the other deep baselines. Our meta-learning resampling strategy, however, was able to produce a 5\% increase over theirs.

\par
\begin{figure}
\begin{minipage}[c]{0.67\textwidth}
    \includegraphics[width=\textwidth]{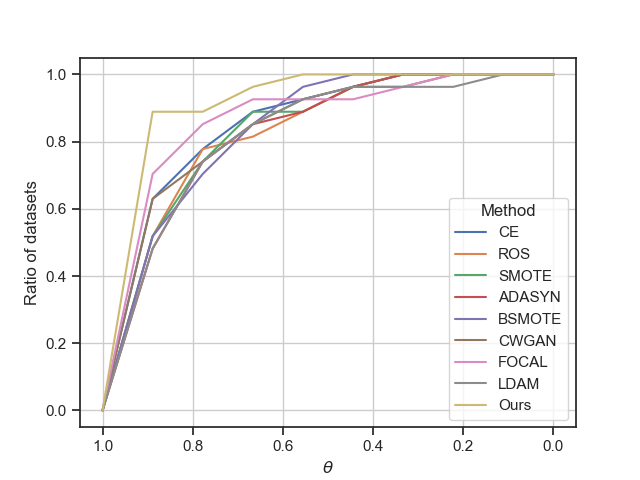}
 \end{minipage}\hfill
 \begin{minipage}[c]{0.3\textwidth}
       \caption{A Dolan-More profile of the AUC-PR scores on the Imbalanced-learn benchmark. The x-axis $\theta$ is a varying threshold factor. For each method in the y-axis we plot the ratio out of 27 datasets for which the attained AUC-PR score is above $\theta$ times the best score on each of the datasets. }
        
     \label{fig:dolan_more}
  \end{minipage}
\end{figure}

 To visualize the results of all the methods on the datasets we employ the Dolan-More profile following \cite{anomaly}. In \cref{fig:dolan_more} we plot for every method the ratio of datasets in which the achieved score is greater than a threshold that is a fraction $0 \leq \theta \leq 1$ of the best score on each datasets. When $\theta$ is close to $0$ the score threshold is low and 
 any method cross the threshold on most datasets. As theta tends to $1$ a leading  method suppose to obtain scores which are higher than the threshold on larger fraction of the datasets. We see that indeed the line corresponding to our method is higher over a wide range of thresholds than any of the other methods.

 

\par \textbf{Discussion:}
Our method demonstrates that it can perform better in multiple imbalanced classification scenarios and domains. In this paper, we are mainly focusing on binary classification since this is the most common task in many applications that involve tabular data. We believe that we can also apply our method also to multiclass classification and long-tail class distribution but we leave it to future research. In theses experiments, we showed that our method prevailed with respect to representative baselines on 27 datasets. These findings provide evidence that a meta-learning approach in which the augmentation of new samples is learned explicitly through the neural network weights could be very effective in alleviating the inherent bias of neural networks. The explicit gradient approach also can be viewed as a sort of regularization to upsampling.
If we set $\eta_2=0$ in the second step of our algorithm, we will get a regular upsampling scheme. We are able to obtain a regularization effect when setting $\eta_2 > 0$ in this case instead of feeding the network with repeated minority instances. Now every incoming batch is formed after applying the second stage of the algorithm on minority instances in the previous iteration. Therefore, it is appropriate to duplicate the minority samples at initialization of the algorithm and not add any more because their position is continually changing during the training process through searching the position which minimize the classifiers' loss. 
\par Tabular data is a challenging domain as the features in tabular datasets are often heterogeneous and come from various unrelated sources. These features are often associated with different units and the dependencies between features and correlation to the target variable could be highly complicated. The main disadvantage of our algorithm is that it has been demonstrated only on shallow MLP classifiers that capture dependencies on a broad range of simple datasets. In order to fully integrate our method with deeper models that use more sophisticated layers like dropout and batch normalization, more work needs to be done. Another drawback is that by design our meta-training is limited to be co-optimized on a single neural network which can severely limit applications that rely on other classifiers such as tree-based models.

\section{Conclusion}
\label{section:conclusion}
This paper investigates the problem of imbalanced learning, which is a common problem in practical classification tasks. We proposed a novel meta-learning approach that learns the proper augmentation via explicit gradient learning. We tested our methods on synthetic and real datasets and found that they outperformed many widely accepted baselines. We believe that using explicit gradient learning has the potential of improving many imbalanced classification tasks and resulting in improved models for many different applications. 
\par \textbf{Reproducibility Statement:} The full implementation of our method is based on PyTorch and is attached as a supplementary zip file which includes the code, data, and configuration files that we used in our experiments. Our method doesn't require costly computational resources and can be run on Google Collab or as a Jupyter notebook on a modest personal computer.

\bibliography{main}

\begin{thebibliography}{24}
\providecommand{\natexlab}[1]{#1}
\providecommand{\url}[1]{\texttt{#1}}
\expandafter\ifx\csname urlstyle\endcsname\relax
  \providecommand{\doi}[1]{doi: #1}\else
  \providecommand{\doi}{doi: \begingroup \urlstyle{rm}\Url}\fi

\bibitem[Awoyemi et~al.(2017)Awoyemi, Adetunmbi, and
  Oluwadare]{awoyemi2017credit}
J.~O. Awoyemi, A.~O. Adetunmbi, and S.~A. Oluwadare.
\newblock Credit card fraud detection using machine learning techniques: A
  comparative analysis.
\newblock In \emph{2017 International Conference on Computing Networking and
  Informatics (ICCNI)}, pages 1--9. IEEE, 2017.

\bibitem[Ben-Baruch et~al.(2021)Ben-Baruch, Ridnik, Zamir, Noy, Friedman,
  Protter, and Zelnik-Manor]{AsymmetricFocalLoss}
E.~Ben-Baruch, T.~Ridnik, N.~Zamir, A.~Noy, I.~Friedman, M.~Protter, and
  L.~Zelnik-Manor.
\newblock Asymmetric loss for multi-label classification, 2021.

\bibitem[Bregman et~al.(2021)Bregman, Lindenbaum, and Rabin]{bregman2021array}
Y.~Bregman, O.~Lindenbaum, and N.~Rabin.
\newblock Array based earthquakes-explosion discrimination using diffusion
  maps.
\newblock \emph{Pure and Applied Geophysics}, 178\penalty0 (7):\penalty0
  2403--2418, 2021.

\bibitem[Cao et~al.(2019)Cao, Wei, Gaidon, Arechiga, and Ma]{LDAM}
K.~Cao, C.~Wei, A.~Gaidon, N.~Arechiga, and T.~Ma.
\newblock Learning imbalanced datasets with label-distribution-aware margin
  loss, 2019.
\newblock URL \url{https://arxiv.org/abs/1906.07413}.

\bibitem[Chawla et~al.(2002)Chawla, Bowyer, Hall, and Kegelmeyer]{Smote}
N.~V. Chawla, K.~W. Bowyer, L.~O. Hall, and W.~P. Kegelmeyer.
\newblock Smote: Synthetic minority over-sampling technique.
\newblock \emph{Journal of Artificial Intelligence Research}, 16:\penalty0
  321–357, Jun 2002.
\newblock ISSN 1076-9757.
\newblock \doi{10.1613/jair.953}.
\newblock URL \url{http://dx.doi.org/10.1613/jair.953}.

\bibitem[Cui et~al.(2019)Cui, Jia, Lin, Song, and Belongie]{cui2019class}
Y.~Cui, M.~Jia, T.-Y. Lin, Y.~Song, and S.~Belongie.
\newblock Class-balanced loss based on effective number of samples.
\newblock In \emph{Proceedings of the IEEE/CVF Conference on Computer Vision
  and Pattern Recognition}, pages 9268--9277, 2019.

\bibitem[Dablain et~al.(2021)Dablain, Krawczyk, and Chawla]{DeepSmote}
D.~Dablain, B.~Krawczyk, and N.~V. Chawla.
\newblock Deepsmote: Fusing deep learning and smote for imbalanced data, 2021.
\newblock URL \url{https://arxiv.org/abs/2105.02340}.

\bibitem[Engelmann and Lessmann(2020)]{WGAN}
J.~Engelmann and S.~Lessmann.
\newblock Conditional wasserstein gan-based oversampling of tabular data for
  imbalanced learning, 2020.
\newblock URL \url{https://arxiv.org/abs/2008.09202}.

\bibitem[Gordini and Veglio(2017)]{gordini2017customers}
N.~Gordini and V.~Veglio.
\newblock Customers churn prediction and marketing retention strategies. an
  application of support vector machines based on the auc parameter-selection
  technique in b2b e-commerce industry.
\newblock \emph{Industrial Marketing Management}, 62:\penalty0 100--107, 2017.

\bibitem[Han et~al.(2005)Han, Wang, and Mao]{BorderlineSmote}
H.~Han, W.-Y. Wang, and B.-H. Mao.
\newblock Borderline-smote: A new over-sampling method in imbalanced data sets
  learning.
\newblock In D.-S. Huang, X.-P. Zhang, and G.-B. Huang, editors, \emph{Advances
  in Intelligent Computing}, pages 878--887, Berlin, Heidelberg, 2005. Springer
  Berlin Heidelberg.
\newblock ISBN 978-3-540-31902-3.

\bibitem[He et~al.(2008)He, Bai, Garcia, and Li]{Adasyn}
H.~He, Y.~Bai, E.~A. Garcia, and S.~Li.
\newblock Adasyn: Adaptive synthetic sampling approach for imbalanced learning.
\newblock In \emph{2008 IEEE international joint conference on neural networks
  (IEEE world congress on computational intelligence)}, pages 1322--1328. IEEE,
  2008.

\bibitem[Kingma and Ba(2017)]{Adam}
D.~P. Kingma and J.~Ba.
\newblock Adam: A method for stochastic optimization, 2017.

\bibitem[Kohavi et~al.(1996)]{adult}
R.~Kohavi et~al.
\newblock Scaling up the accuracy of naive-bayes classifiers: A decision-tree
  hybrid.
\newblock In \emph{Kdd}, volume~96, pages 202--207, 1996.

\bibitem[Krawczyk(2016)]{challenges}
B.~Krawczyk.
\newblock Learning from imbalanced data: Open challenges and future directions.
\newblock \emph{Progress in Artificial Intelligence}, 5, 04 2016.
\newblock \doi{10.1007/s13748-016-0094-0}.

\bibitem[Lema{{\^i}}tre et~al.(2017)Lema{{\^i}}tre, Nogueira, and
  Aridas]{imblearn}
G.~Lema{{\^i}}tre, F.~Nogueira, and C.~K. Aridas.
\newblock Imbalanced-learn: A python toolbox to tackle the curse of imbalanced
  datasets in machine learning.
\newblock \emph{Journal of Machine Learning Research}, 18\penalty0
  (17):\penalty0 1--5, 2017.
\newblock URL \url{http://jmlr.org/papers/v18/16-365.html}.

\bibitem[Lin et~al.(2018)Lin, Goyal, Girshick, He, and Dollár]{FocalLoss}
T.-Y. Lin, P.~Goyal, R.~Girshick, K.~He, and P.~Dollár.
\newblock Focal loss for dense object detection, 2018.

\bibitem[Lindenbaum et~al.(2018)Lindenbaum, Stanley~III, Wolf, and
  Krishnaswamy]{Sugar}
O.~Lindenbaum, J.~S. Stanley~III, G.~Wolf, and S.~Krishnaswamy.
\newblock Geometry-based data generation.
\newblock \emph{arXiv preprint arXiv:1802.04927}, 2018.

\bibitem[Liu et~al.(2020)Liu, Wei, Jiang, Cao, Bian, and Chang]{mesa}
Z.~Liu, P.~Wei, J.~Jiang, W.~Cao, J.~Bian, and Y.~Chang.
\newblock Mesa: Boost ensemble imbalanced learning with meta-sampler.
\newblock \emph{Advances in Neural Information Processing Systems}, 33, 2020.

\bibitem[Nguyen et~al.(2017)Nguyen, Le, Vu, and Phung]{D2GAN}
T.~D. Nguyen, T.~Le, H.~Vu, and D.~Phung.
\newblock Dual discriminator generative adversarial nets, 2017.
\newblock URL \url{https://arxiv.org/abs/1709.03831}.

\bibitem[Odena et~al.(2017)Odena, Olah, and Shlens]{ACGAN}
A.~Odena, C.~Olah, and J.~Shlens.
\newblock Conditional image synthesis with auxiliary classifier gans.
\newblock In \emph{International conference on machine learning}, pages
  2642--2651. PMLR, 2017.

\bibitem[Paszke et~al.(2019)Paszke, Gross, Massa, Lerer, Bradbury, Chanan,
  Killeen, Lin, Gimelshein, Antiga, Desmaison, Kopf, Yang, DeVito, Raison,
  Tejani, Chilamkurthy, Steiner, Fang, Bai, and Chintala]{PyTorch}
A.~Paszke, S.~Gross, F.~Massa, A.~Lerer, J.~Bradbury, G.~Chanan, T.~Killeen,
  Z.~Lin, N.~Gimelshein, L.~Antiga, A.~Desmaison, A.~Kopf, E.~Yang, Z.~DeVito,
  M.~Raison, A.~Tejani, S.~Chilamkurthy, B.~Steiner, L.~Fang, J.~Bai, and
  S.~Chintala.
\newblock Pytorch: An imperative style, high-performance deep learning library.
\newblock In H.~Wallach, H.~Larochelle, A.~Beygelzimer, F.~d\textquotesingle
  Alch\'{e}-Buc, E.~Fox, and R.~Garnett, editors, \emph{Advances in Neural
  Information Processing Systems 32}, pages 8024--8035. Curran Associates,
  Inc., 2019.
\newblock URL
  \url{http://papers.neurips.cc/paper/9015-pytorch-an-imperative-style-high-performance-deep-learning-library.pdf}.

\bibitem[Rabin et~al.(2016)Rabin, Bregman, Lindenbaum, Ben-Horin, and
  Averbuch]{rabin2016earthquake}
N.~Rabin, Y.~Bregman, O.~Lindenbaum, Y.~Ben-Horin, and A.~Averbuch.
\newblock Earthquake-explosion discrimination using diffusion maps.
\newblock \emph{Geophysical Journal International}, 207\penalty0 (3):\penalty0
  1484--1492, 2016.

\bibitem[Shenkar and Wolf(2021)]{anomaly}
T.~Shenkar and L.~Wolf.
\newblock Anomaly detection for tabular data with internal contrastive
  learning.
\newblock In \emph{International Conference on Learning Representations}, 2021.

\bibitem[Shu et~al.(2019)Shu, Xie, Yi, Zhao, Zhou, Xu, and Meng]{metaweightnet}
J.~Shu, Q.~Xie, L.~Yi, Q.~Zhao, S.~Zhou, Z.~Xu, and D.~Meng.
\newblock Meta-weight-net: Learning an explicit mapping for sample weighting,
  2019.

\end{thebibliography}

\section*{Checklist}

The checklist follows the references.  Please
read the checklist guidelines carefully for information on how to answer these
questions.  For each question, change the default \answerTODO{} to \answerYes{},
\answerNo{}, or \answerNA{}.  You are strongly encouraged to include a {\bf
justification to your answer}, either by referencing the appropriate section of
your paper or providing a brief inline description.  For example:
\begin{itemize}
  \item Did you include the license to the code and datasets? \answerYes{See Section~\ref{gen_inst}.}
  \item Did you include the license to the code and datasets? \answerNo{The code and the data are proprietary.}
  \item Did you include the license to the code and datasets? \answerNA{}
\end{itemize}
Please do not modify the questions and only use the provided macros for your
answers.  Note that the Checklist section does not count towards the page
limit.  In your paper, please delete this instructions block and only keep the
Checklist section heading above along with the questions/answers below.


\begin{enumerate}

\item For all authors...
\begin{enumerate}
  \item Do the main claims made in the abstract and introduction accurately reflect the paper's contributions and scope?
    \answerYes{}
  \item Did you describe the limitations of your work?
    \answerYes{}
  \item Did you discuss any potential negative societal impacts of your work?
    \answerNA{}{}
  \item Have you read the ethics review guidelines and ensured that your paper conforms to them?
    \answerYes{}{}
\end{enumerate}

\item If you are including theoretical results...
\begin{enumerate}
  \item Did you state the full set of assumptions of all theoretical results?
    \answerNA{}
        \item Did you include complete proofs of all theoretical results?
    \answerNA{}
\end{enumerate}

\item If you ran experiments...
\begin{enumerate}
  \item Did you include the code, data, and instructions needed to reproduce the main experimental results (either in the supplemental material or as a URL)?
    \answerYes{}
  \item Did you specify all the training details (e.g., data splits, hyperparameters, how they were chosen)?
    \answerYes{In the experiments sections}
        \item Did you report error bars (e.g., with respect to the random seed after running experiments multiple times)?
    \answerYes{The std values on validation folds was reported}
        \item Did you include the total amount of compute and the type of resources used (e.g., type of GPUs, internal cluster, or cloud provider)?
    \answerYes{}
\end{enumerate}

\item If you are using existing assets (e.g., code, data, models) or curating/releasing new assets...
\begin{enumerate}
  \item If your work uses existing assets, did you cite the creators?
    \answerYes{}
  \item Did you mention the license of the assets?
    \answerNA{}
  \item Did you include any new assets either in the supplemental material or as a URL?
    \answerYes{see \ref{section:conclusion}}
  \item Did you discuss whether and how consent was obtained from people whose data you're using/curating?
    \answerNo{The code and the data are
proprietary.}
  \item Did you discuss whether the data you are using/curating contains personally identifiable information or offensive content?
   \answerNo{The data is publicly available on the internet with no privacy restrictions}
\end{enumerate}

\item If you used crowdsourcing or conducted research with human subjects...
\begin{enumerate}
  \item Did you include the full text of instructions given to participants and screenshots, if applicable?
    \answerNA{}
  \item Did you describe any potential participant risks, with links to Institutional Review Board (IRB) approvals, if applicable?
    \answerNA{}
  \item Did you include the estimated hourly wage paid to participants and the total amount spent on participant compensation?
    \answerNA{}
\end{enumerate}

\end{enumerate}

\appendix
\section{A: Additional Sine Wave results}
\label{appendix:a}
\begin{tabular}{c}
    \adjustimage{width=\textwidth, height=0.15\height, valign=M}{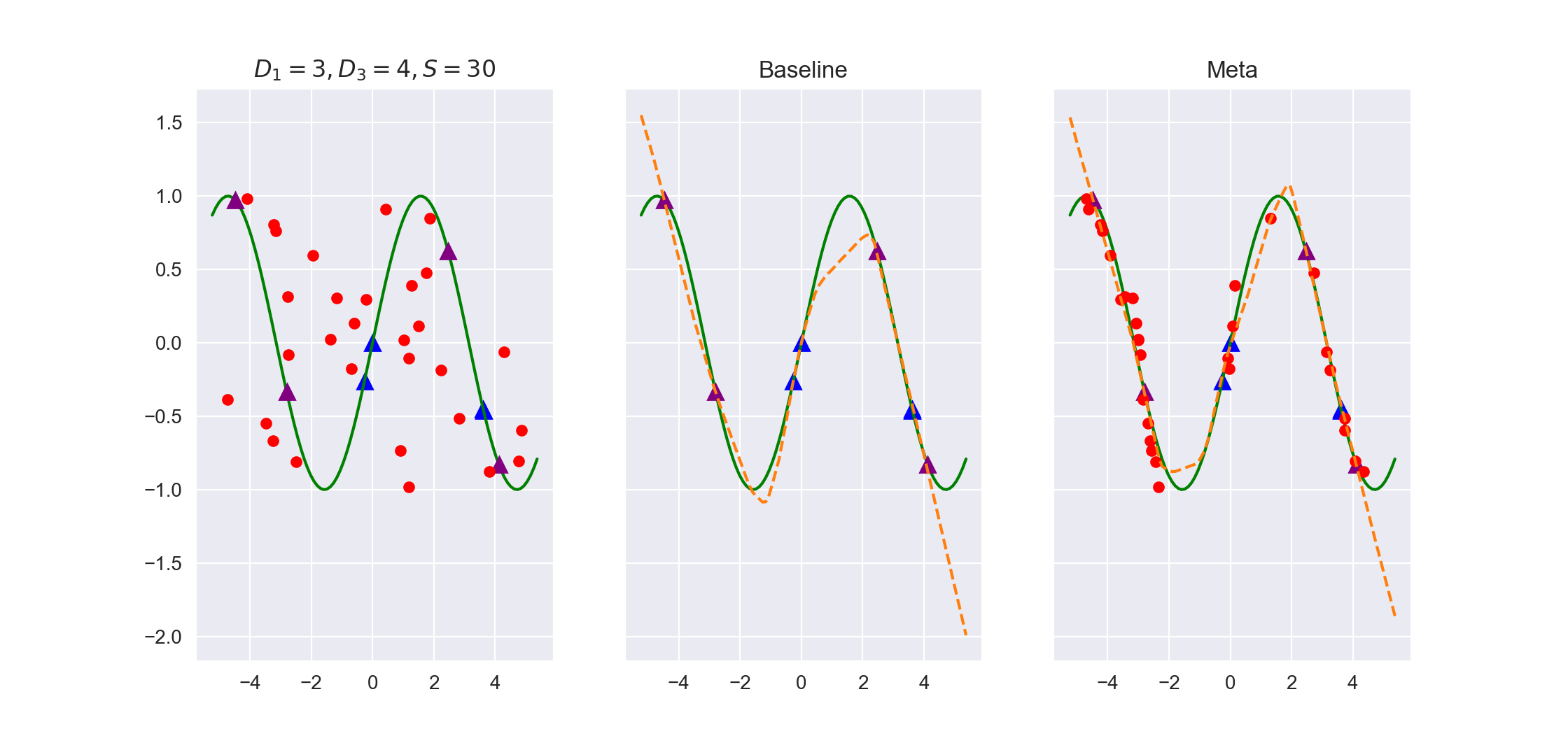} \\
  \adjustimage{width=\textwidth, height=0.15\height, valign=M}{figures/sine/sin1.png} \\
  \adjustimage{width=\textwidth, height=0.15\height, valign=M}{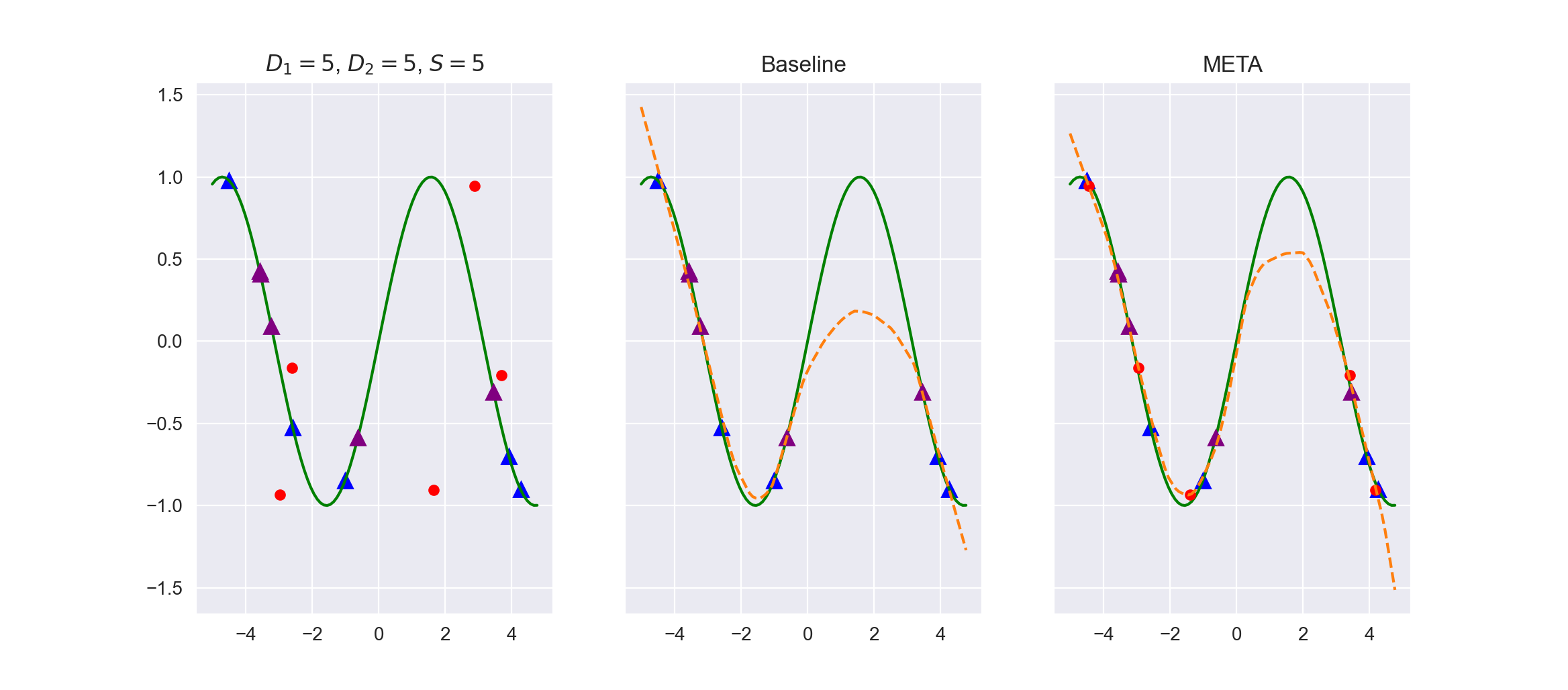} \\
  \adjustimage{width=\textwidth,height=0.15\height, valign=M}{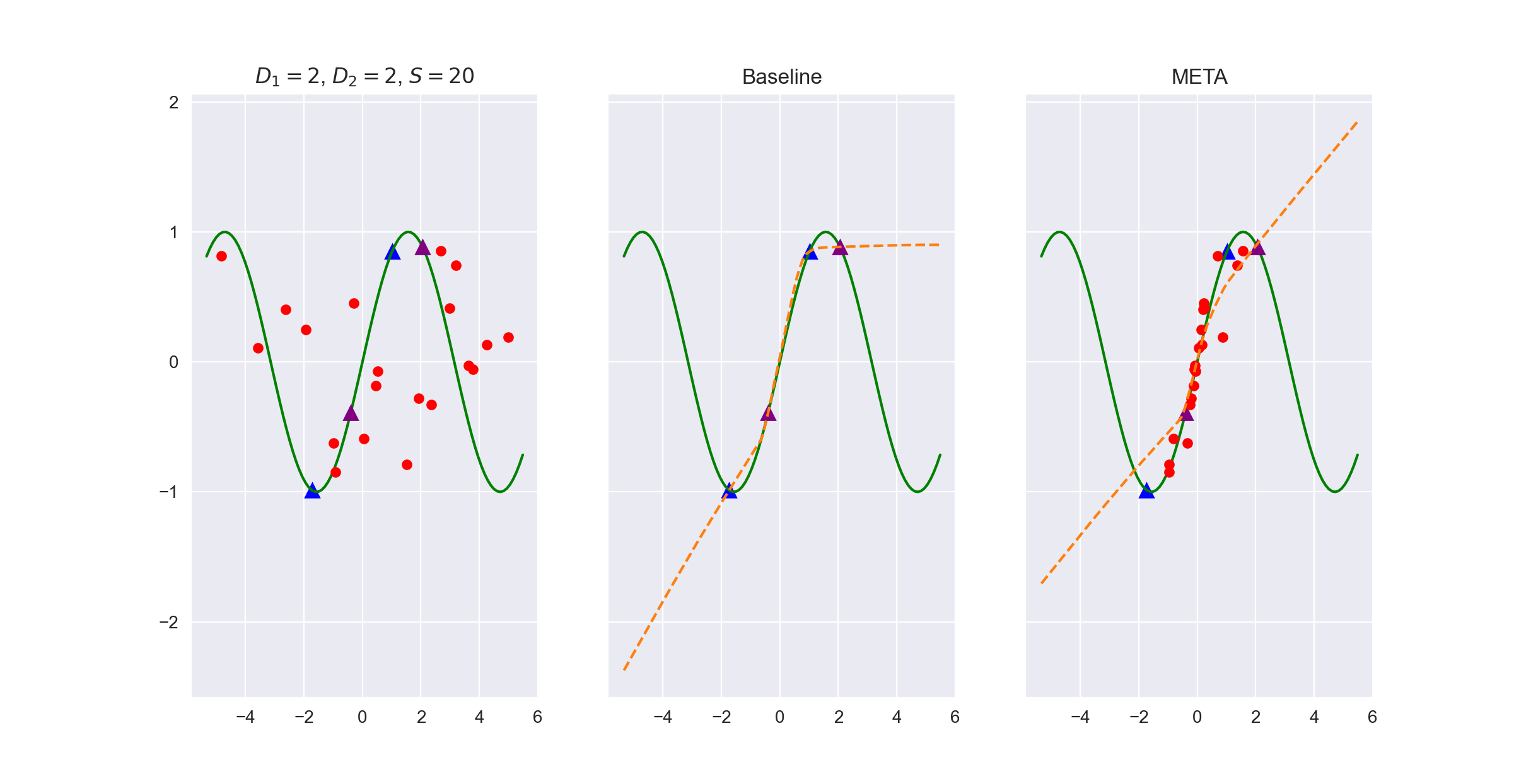}\\

\end{tabular}


\section{B: Imblearn Dataset Statistics}
\label{appendix:b}

In Tab.\ref{table:datasets} we present the full characteristics of the datasets integrated within the  \cite{imblearn} package.

\begin{table}[ht]
\centering
\adjustbox{width=0.8\textwidth}{

\begin{tabular}{@{}lllll@{}}
\toprule
ID & Name             & \#F & \#S     & Imbalance Ratio \\ \midrule
1  & ecoli            & 7   & 336     & 8.6:1           \\
2  & optical\_digits  & 64  & 5,620   & 9.1:1           \\
3  & satimage         & 36  & 6,435   & 9.3:1           \\
4  & pen\_digits      & 16  & 10,992  & 9.4:1           \\
5  & abalone          & 10  & 4,177   & 9.7:1           \\
6  & sick\_euthyroid  & 42  & 3,163   & 9.8:1           \\
7  & spectrometer     & 93  & 531     & 11:1            \\
8  & car\_eval\_34    & 21  & 1,728   & 12:1            \\
9  & isolet           & 617 & 7,797   & 12:1            \\
10 & us\_crime        & 100 & 1,994   & 12:1            \\
11 & yeast\_ml8       & 103 & 2,417   & 13:1            \\
12 & scene            & 294 & 2,407   & 13:1            \\
13 & libras\_move     & 90  & 360     & 14:1            \\
14 & thyroid\_sick    & 52  & 3,772   & 15:1            \\
15 & coil\_2000       & 85  & 9,822   & 16:1            \\
16 & arrhythmia       & 278 & 452     & 17:1            \\
17 & solar\_flare\_m0 & 32  & 1,389   & 19:1            \\
18 & oil              & 49  & 937     & 22:1            \\
19 & car\_eval\_4     & 21  & 1,728   & 26:1            \\
20 & wine\_quality    & 11  & 4,898   & 26:1            \\
21 & letter\_img      & 16  & 20,000  & 26:1            \\
22 & yeast\_me2       & 8   & 1,484   & 28:1            \\
23 & webpage          & 300 & 34,780  & 33:1            \\
24 & ozone\_level     & 72  & 2,536   & 34:1            \\
25 & mammography      & 6   & 11,183  & 42:1            \\
26 & protein\_homo    & 74  & 145,751 & 111:1           \\
27 & abalone\_19      & 10  & 4,177   & 130:1           \\ \bottomrule
\end{tabular}
}
\caption{The features dimension, number of samples and the imbalance ratio for the datasets used in the second set of experimetns}
\label{table:datasets}
\end{table}
\section{C}

\label{appendix:c}
\begin{table}[]
\caption{Convergence verification of the trained model with the synthetic augmentation on \textbf{Adult} with IR $50:1$.}
\begin{tabular}{c}

\adjustimage{width=0.3\textwidth, height=0.3\height, valign=M}{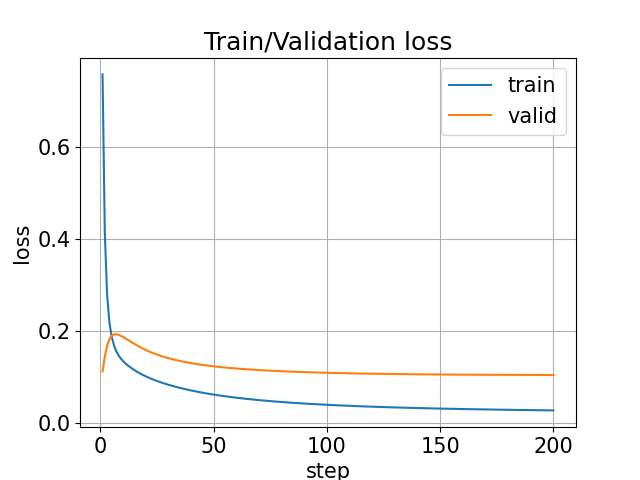} 
\adjustimage{width=0.3\textwidth, height=0.3\height, valign=M}{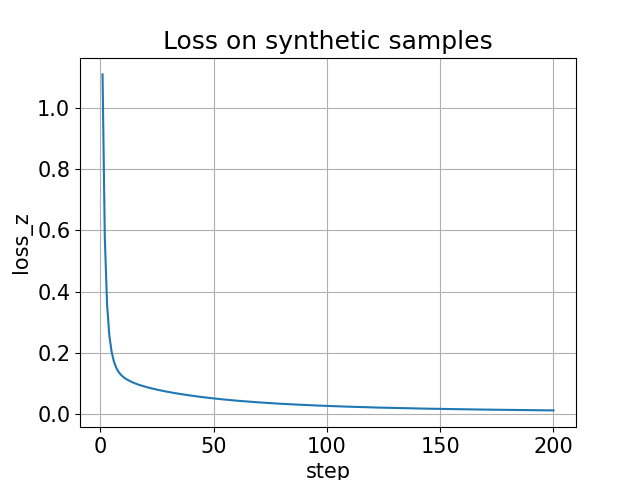} 
\adjustimage{width=0.3\textwidth, height=0.3\height, valign=M}{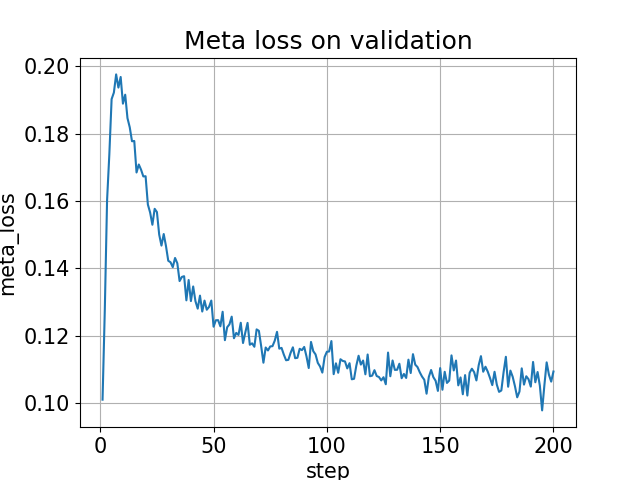} 
\end{tabular}
 
\end{table}   

\end{document}